\documentclass[conference]{IEEEtran}
\IEEEoverridecommandlockouts
\usepackage{cite}
\usepackage{amsmath,amssymb,amsfonts}
\usepackage{algorithmic}
\usepackage{graphicx}
\usepackage{textcomp}
\usepackage{xcolor}

\usepackage{multirow}
\usepackage{hyperref}
\usepackage{tcolorbox}
\usepackage{orcidlink}
\usepackage{enumitem}

\def\BibTeX{{\rm B\kern-.05em{\sc i\kern-.025em b}\kern-.08em
    T\kern-.1667em\lower.7ex\hbox{E}\kern-.125emX}}

\begin{document}

\title{Requirement Engineering Challenges \\ for  AI-intense Systems Development
\thanks{This project has received funding from the European Union’s Horizon 2020 research and innovation program under grant agreement No 957197 and from the European Union’s Horizon 2020 research and innovation programme under the Marie Skłodowska-Curie grant agreement No 860410.}
}

\author{
\IEEEauthorblockN{Hans-Martin Heyn\IEEEauthorrefmark{1}\IEEEauthorrefmark{4}\orcidlink{0000-0002-2427-6875},
Eric Knauss\IEEEauthorrefmark{1}\orcidlink{0000-0002-6631-872X},
Amna Pir Muhammad\IEEEauthorrefmark{1}\orcidlink{0000-0001-8328-4149},Olof Eriksson\IEEEauthorrefmark{3}
\\
Jennifer Linder\IEEEauthorrefmark{1},
Padmini Subbiah\IEEEauthorrefmark{1},
Shameer Kumar Pradhan\IEEEauthorrefmark{1},
Sagar Tungal\IEEEauthorrefmark{1}}

\IEEEauthorblockA{\IEEEauthorrefmark{1}\textit{Department of Computer Science and Engineering} \\
\textit{Chalmers $\mid$ University of Gothenburg, Sweden}\\
\IEEEauthorrefmark{3}\textit{Veoneer Sweden AB} \\}
\IEEEauthorblockA{\IEEEauthorrefmark{4}Corresponding author, Hans-Martin.Heyn@gu.se}
}

\maketitle

\begin{abstract}
Availability of powerful computation and communication technology as well as advances in artificial intelligence enable a new generation of complex, AI-intense systems and applications.
Such systems and applications promise exciting improvements on a societal level, yet they also bring with them new challenges for their development. 
In this paper we argue that significant challenges relate to defining and ensuring behaviour and quality attributes of such systems and applications.
We specifically derive four challenge areas from relevant use cases of complex, AI-intense systems and applications related to industry, transportation, and home automation: understanding, determining, and specifying 
(i) contextual definitions and requirements, 
(ii) data attributes and requirements,  
(iii) performance definition and monitoring, and 
(iv) the impact of  human factors on system acceptance and success.
Solving these challenges will imply process support that integrates new requirements engineering methods into development approaches for complex, AI-intense systems and applications. 
We present these challenges in detail and propose a research roadmap. 
\end{abstract}

\vspace{\baselineskip}


% !TEX root = main.tex
\section{Introduction}

%\emph{Why is the topic of interest?}\par
The availability of cheap computational power and communication technology as well as innovations in artificial intelligence (AI) allows for increasingly complex applications with fundamental societal impact, for example in the areas of industry, transportation, and in our homes. The increasing functionality of future technical systems is accompanied by a higher design and management complexity. A purely manual configuration of the upcoming complex devices and networks will become more and more difficult. Furthermore, current systems engineering approaches for building such systems start to fail and cannot be applied to find resource-efficient solutions to these applications \cite{Bosch2020}. The amounts of data collected to be processed are extremely large, the computational power required is extreme, and thus requires large amounts of energy, and the algorithms are too complex and cannot deliver solutions within the tight time constraints. Even describing requirements and constraints such as security, privacy, or robustness for such systems, and ensuring that these are fulfilled becomes a critical challenge and threatens the deployment of such futuristic applications to society. A fundamental challenge for developing such applications and systems relates to the fact that AI components do not behave as conventional components \cite{Horkoff2019}. In particular, they relate to a new and different set of quality attributes \cite{Horkoff2019} and they imply a certain level of uncertainty, thus forcing for example a shift from functional safety to the idea of safety of the intended functionality \cite{SOTIF,Borg2019,Martin2017}. For example, the function \emph{object recognition} implies to identify objects in a camera feed with a certain accuracy. As long as this (promised) accuracy is matched, such an AI-enabled component is thought to fulfil its requirements. However, this can even mean that at times, existing objects are not identified or identifications are reported where no objects were present.In addition, the accuracy may depend on context (especially how good that context matches the training data) as well as on the quality of input data.
In order to build reliable systems of such components, we argue that these specifics must be considered when engineering requirements of \emph{AI-intense} systems (as part of the systems engineering). We call the system \emph{AI-intense} if the functionality of the AI fundamentally determines the overall functionality of the system. \par
%\emph{What is the background on the previous solutions if any?}
%\emph{What is the background on potential solutions?}
It is our ambition to address this gap in the scope of two research projects: Very Efficient Deep Learning in Internet of Things (VEDLIoT) and Supporting the Interaction of Humans and Automated Vehicles: Preparing for the Environment of Tomorrow (SHAPE-IT)\footnote{See \href{https://vedliot.eu}{https://vedliot.eu} and \href{https://www.shape-it.eu}{https://www.shape-it.eu}.}.
VEDLIoT is concerned with a technological platform for AI-intense systems and allows us to investigate appropriate decomposition of requirements and architecture. Our reasoning in this paper is based on VEDLIoT use cases from three different domains: industry, transportation, and smart home.
SHAPE-IT focuses on the transportation domain, but specifically focuses on the interaction of AI-intense automatic vehicles (AV) with users and other traffic participants.
%\emph{what was/will be attempted in the present effort (research project)?}
Thus, these projects complement each other with respect to needs for new RE methods and align on the research objective for this paper: \par
\textbf{Research objective:} Determine relevant challenges related to engineering requirements of AI-intense systems.

Based on three use cases of distributed deep learning, we describe challenges related to systems engineering. 
In particular, these relate to ensuring certain behaviour and quality attributes of the overall system and its applications, i.e. the quality in use.
We argue that this is a conceptual and methodological problem related to requirements engineering. \par

%\emph{What will be presented in this paper?}
\textbf{Contribution:} Our ongoing work suggests challenges in particular with respect to the following areas:
\begin{itemize}
    \item {Understanding, determining, and specifying the context of operation for complex, AI-intense systems and applications.}
    \item {Understanding, determining, and specifying quality of data related to the design and operation of complex, AI-intense systems and applications. This includes data required for training of the AI, because the training process eventually creates the functionality of the AI system.}
    \item {Deriving operationalisations from context and data quality that allow to test complex, AI-intense systems and applications as well as to monitor their performance during runtime.}
    \item Including human factors knowledge in the design of AI-intense systems.
\end{itemize}
While solutions to these challenges will help to guarantee functionality and behaviour as well as quality effectively, it will also be crucial to integrate such solutions into development approaches. Managing knowledge related to the above (context, data quality, operationalisations) within the development life cycle of complex, AI-intense systems and applications is therefore a cross-cutting concern. This paper contributes our reflections on a suitable research roadmap for these challenges. \par

%Section \ref{sec:use-cases} presents our use cases under analysis.
%Based on these, we describe the identified research problems in Section \ref{sec:research-problems}. We present our conclusions and outlook on future work in Section \ref{sec:conclusion}.

% !TEX root = main.tex
\section{Example use cases}
\label{sec:use-cases}

This section presents three relevant use cases of complex, AI-intense systems. The use cases represent different fields of application of AI systems. All use cases have in common that an AI is the core of the decision making process for the application. Our aim is to identify universal challenges when applying AI in a complex system.

\subsection{Use Case 1: Automation industry / Fault detection} \par
In industrial Internet of things (IoT) applications, deep learning promises to increase safety by allowing to automatically detect and resolve critical situations. In such applications, decisions must typically be made within a few milliseconds in order to meet safety and control requirements. Communication steps from the sensor via a decision making unit to the actuator might violate timing requirements. To overcome this problem, trained machine learning models at the sensor node could lower the overall response time of the system and allow early fault detection. \par
\begin{figure}
    \centering
    \includegraphics[width=0.6\linewidth]{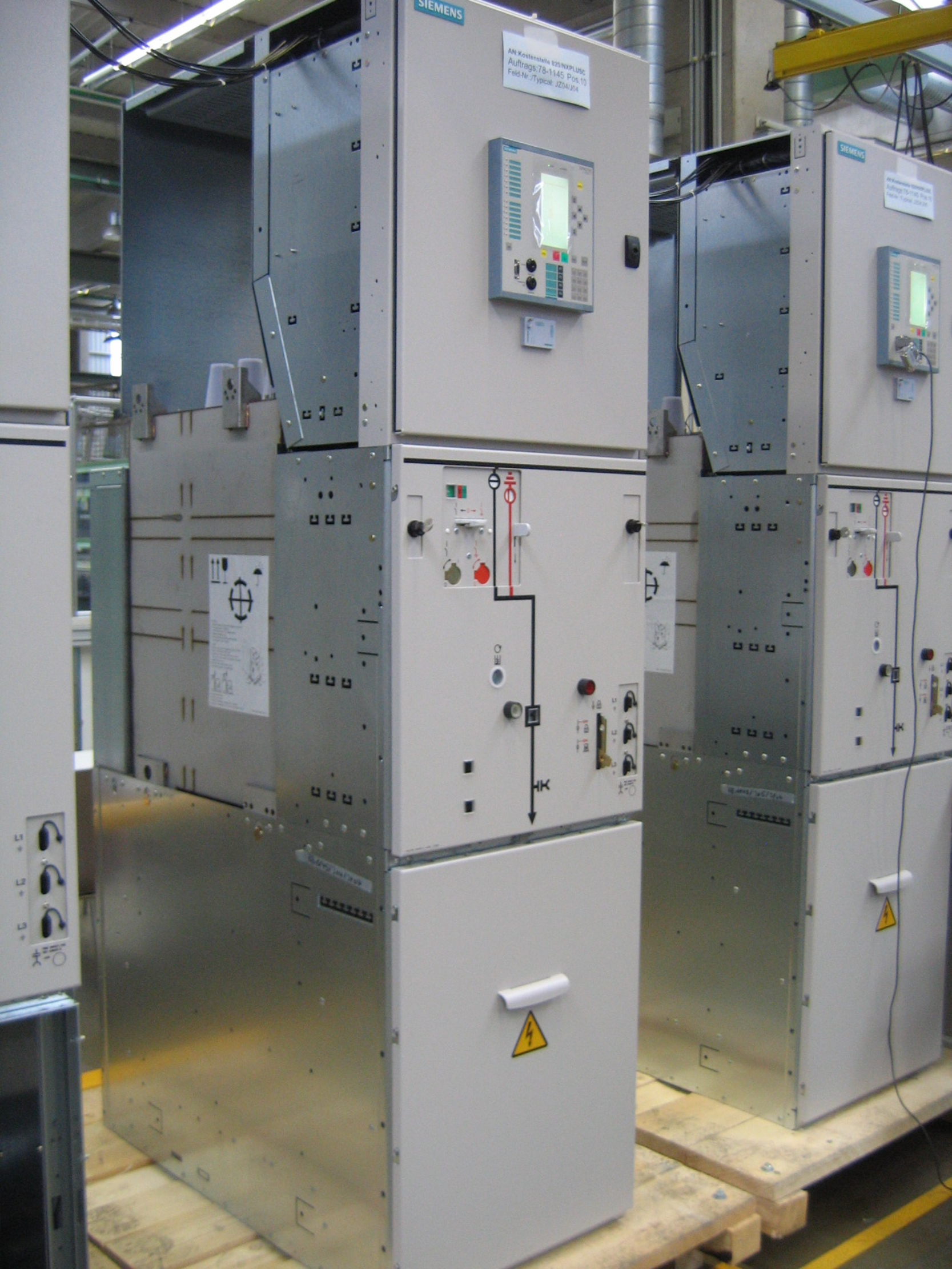}
    \caption{A typical electric switchboard (Wikimedia,\\ \url{https://commons.wikimedia.org/wiki/File:Medium_voltage_panel.jpg};\\ Retrieved 05/01/2021)}
    \label{fig:01_switchboard}
\end{figure}
In use case 1, pattern recognition with a machine learning model enables fast arc detection for distribution switchboards as shown in Figure~\ref{fig:01_switchboard}. An electric arc in the switchboard typically occurs during a switching fault, and it is crucial to trigger a circuit breaker as fast as possible to avoid damage to electric equipment or persons in close vicinity. The characteristics of the arc depend on the load type, load characteristics, and the impedance of the connected infrastructure. \par
The sensor data are collected from different types of sensors (IR-camera, temperature sensor, magnetic field sensor) combined in a single housing with some processing capability. The sensor node will employ pattern recognition with deep learning for fast fault detection. The machine learning model for fault detection will be trained with appropriate training data based on the typical characteristics of the arc. \par
In case the characteristics of the arc change, e.g., due to ageing of components, changes in loads, or by deploying an identical sensor node in another switchboard with different loads, it cannot be guaranteed that the trained machine learning model is still able to detect faults with sufficient performance. The context in which the machine learning model operates has been changed. But because the neural network has been trained and tested with a set of training and test data until a desired performance for all at that time known arc characteristics have been achieved, it can only perform in that given context. By selecting the training and test data, the context in which the sensor node will perform has been fixed.

\subsection{Use Case 2: Automotive industry / ADAS} \par
This second use case concerns automotive AI and specifically automatic emergency braking (AEB) and AEB for mitigating collisions with pedestrians. AEB is a fundamental function for an advanced driver assistance system (ADAS). Emergency braking is a safety critical function and strict functional safety requirements are imposed on the system. A typical functional safety requirement for AEB is: \par
\begin{figure}
    \centering
    \includegraphics[width=\linewidth]{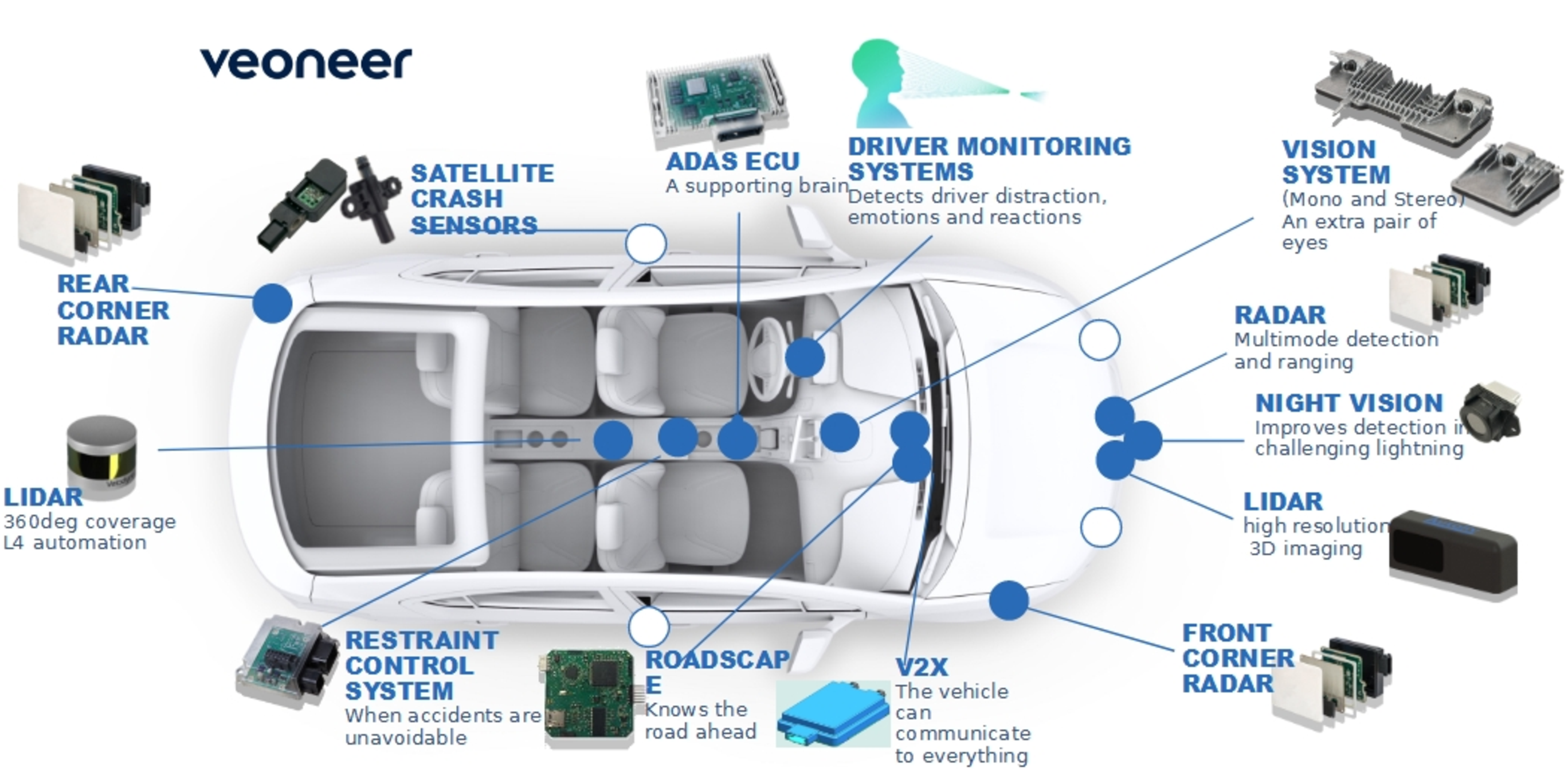}
    \caption{A typical sensor configuration in a modern passenger vehicle. Courtesy of Veoneer Sweden.}
    \label{fig:02_VEDLIOT_Automotive}
\end{figure}

\begin{center}
    \emph{Unwanted emergency braking shall not occur.}
\end{center}
As illustrated in Figure~\ref{fig:02_VEDLIOT_Automotive}, a modern passenger car contains a vast variety of different sensors with different technologies and different safety ratings. Some sensor nodes, especially those with vision based cameras, contain trained machine learning models close to the sensor for object detection and decision making. Today, each sensor normally serves a clear purpose, {has a clear set of safety requirements assigned to it}, and provides data to a fixed set of functions. In the future this setup is up to a challenge and in our opinion the following are the most important ones:
\begin{enumerate}
    \item In future configurations, the sensors could adapt to the current driving situation. For example a vision camera might be useless in darkness or heavy fog, and its processing power can instead increase precision of the radar. 
    \item With the introduction of remote software downloads in the automotive industries, new functions might be added after the vehicle has been deployed. The set of functions a sensor provides data to might change during the lifetime of the vehicle.
    \item The context in which the vehicle can operate can change drastically and unpredictably e.g. due to a malfunction or accident of the vehicle. However, especially for highly automated driving, the vehicle (and its sensors) must still be able to ensure a safe stop of the vehicle.
    \item The performance of the sensors and AI-intense systems need to be monitored in real-time in order to detect faults early and to mitigate potentially dangerous driving situation (safety of the intended function \cite{SOTIF}).
    \item ADAS functionality increasingly relies on data obtained from the cloud (e.g. traffic or weather information) and it might be desirable to move parts of the decision process to the cloud.
\end{enumerate}

\subsection{Use Case 3: Smart Home} \par
Artificial intelligence enables many complex smart home applications. Examples are smart doors that contain face recognition, smart assistants which employ natural language processing (NLP), or intelligent fitness coaches with gesture detection. Another attribute of a smart home is that the sensor, edge processing unit, and actuator nodes are distributed over a local area and permanently connected to cloud services. \par
\begin{figure}
    \centering
    \includegraphics[width=0.35\linewidth]{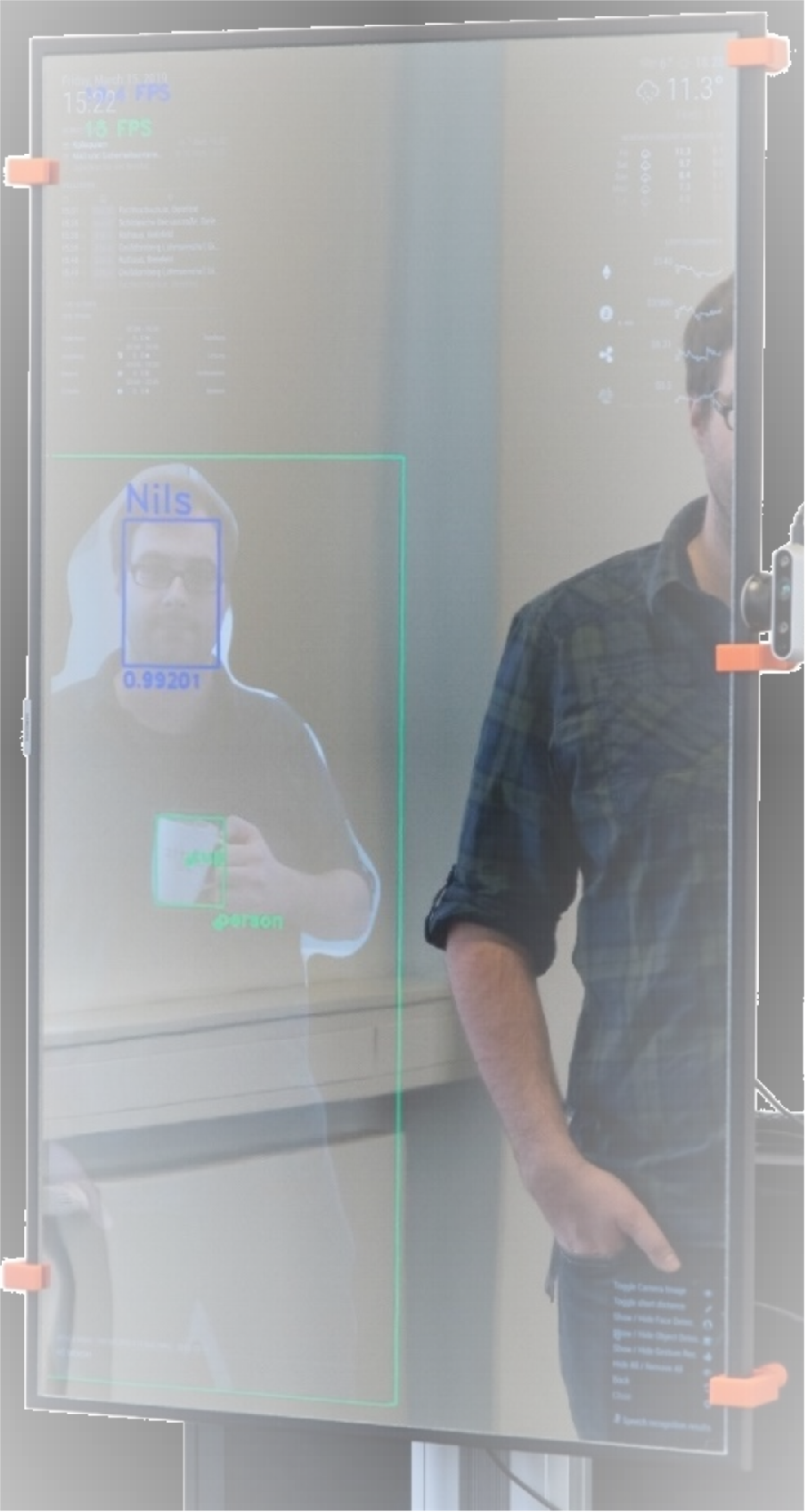}
    \caption{A smart mirror as example for a smart home application \cite{Salami2020}}
    \label{fig:03_smartmirrow}
\end{figure}
A smart mirror as shown in Figure~\ref{fig:03_smartmirrow} will be used as showcase object in VEDLIoT. A camera, microphone, and display coated with a semi-transparent foil provide, in combination with several neural networks, face recognition, object detection, gesture detection and speech recognition. \par
The most critical aspect of personal systems is data privacy and security of the system. Strict requirements will be imposed to ensure the integrity of privacy. Most privacy-critical data processing shall happen close to the data collection point (edge computation) \cite{Salami2020}. Furthermore, in contrast to industrial applications, it can be difficult to exactly frame the context in which a private consumer product is being used. In case of the smart mirror, it could be used in a private home, or in a public area such as a hospital or an open public space such as a train station. This will have affects on the expected input data quality to the trained learning models (background noise in public areas for example) and requirements on data integrity (such as different privacy concerns). \par
% !TEX root = main.tex
\section{Identified research problems}
\label{sec:research-problems}

VEDLIoT aims at building tools and methodologies for supporting AI development for the presented use cases. The use cases are provided by industry partners, and additional use case might enter the project at a later stage. During discussions with the industry partners and academic partners involved in VEDLIoT we observed several reoccurring problems being mentioned. In an internal brainstorming workshop between the industry partner providing Use Case 2 and the University of Gothenburg, we categorised the problems in the following four problem areas:

\begin{description}
    \item[PA~1:] Contextual definition and requirements; The ability of ensuring a desired system behaviour requires that the context, in which the machine learning model is deployed, is clearly defined.
    \item[PA~2:] Data attributes and requirements; The context, and especially the ability to guarantee certain system quality attributes such as safety and robustness, on the other hand will impose non-functional requirements which will lead to requirements of the data in use. 
    \item[PA~3:] Performance metrics, reproducibility, comparability and real-time monitoring of trained machine learning models; To achieve continuous improvement of the system and to enable the system to react on situations where functionality and quality attributes can no longer be satisfied, performance monitoring and reporting needs to be established in the given context. 
    \item[PA~4:] Human Factors; For the success of the system, human factors must be considered - will humans accept decisions of the automated system? Will they react accordingly? Will this affect the performance of the system in use?
\end{description}

\begin{figure}
    \centering
    \includegraphics[width=\linewidth]{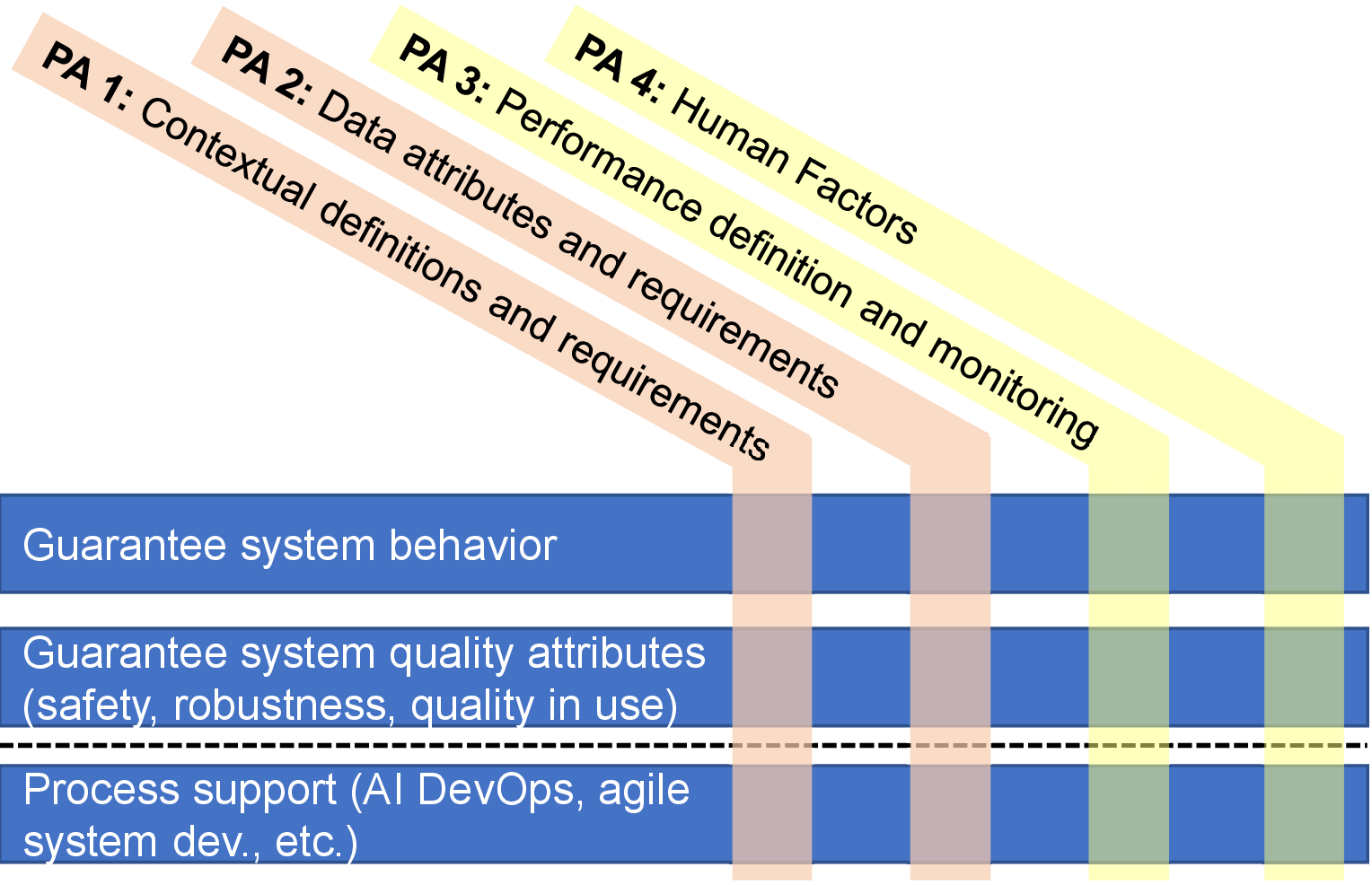}
    \caption{The development of complex, AI-intense systems implies the need of certain abilities (blue boxes) that depend on solutions for challenges in four areas (red/yellow boxes). We argue that one has to find solutions for the red challenge areas before approaching the yellow challenge area.}
    \label{fig:04_challenges}
\end{figure}

Figure~\ref{fig:04_challenges} illustrates that all four problems are interconnected with each other. The first three problem areas relate strongly to the scope of the VEDLIoT project and are complemented with a less system focused problem area related to the SHAPE-IT project. \par
In addition, a cross-cutting aspect relates to process support for modern system development approaches and the success of solutions within each problem area depend on good integration into engineering practice.

Given the dependencies between the problem areas, we focus on the first two problem areas, because good conceptual approaches for managing context and data requirements will support identifying solutions in the other problem areas.

\subsection{\textbf{PA 1}: Contextual definitions and requirements}
In the automotive use case, the problem of contextual definitions is probably most obvious: Today, an ADAS cannot operate in any given driving situation. The system is designed and tested to perform safely only in a priori defined conditions. According to SAE \cite{j3016}, these operating conditions define the operational design domain (ODD). A significant problem for the development of more automated vehicles is a lack of a common definition for the ODD of a vehicle \cite{Koopman2019,Gyllenhammar2020}. Still, the original equipment manufacturer (OEM) must be able to guarantee a certain system behaviour and especially safety attributes.\par
If deep learning is used to enable complex object (people, obstacles) and pattern (road markings) detection, the ODD, as a way to describe the context in which the vehicle will operate, will govern the required performance of the trained machine learning model. The problem of contextual definition can even be abstracted beyond the automotive use case to other applications of machine learning:\par
A trained machine learning model cannot be placed into another context without appropriate new training and testing. The context can also change slowly over time. To preserve the ability of ensuring the system's behaviour, significant more transparency about the entire life-cycle of a machine learning model will be required. It must be shown that the context, in which a machine learning model operates, is suitable. A starting point towards more transparency for machine learning models are model cards \cite{Mitchell2019}.\par
The importance of proper context definition for machine learning becomes apparent in the \emph{no-free-lunch-theorems} \cite{Wolpert1996}. In brief they state that over \emph{all} data-generating distributions, every machine learning algorithm will perform equally poor when confronted with previously unobserved data. It is necessary to make assumptions on the probability distributions that the trained model is expected to encounter in the application. Those assumptions, or beliefs, can be explicit by directly stating assumed probability distributions over parameters of the model. They can also be implicit by choosing learning algorithm that are biased towards choosing some class of function over another \cite{Goodfellow2016}. A link obviously exists between the context of an application and the expected data attributes.\par

\begin{tcolorbox}[title=Research Questions: Contextual definitions]
\begin{description}
    \item[RQ 1-1:] What challenges arise when deriving contextual definitions and requirements from use cases?
    \item[RQ 1-2:] Which practices would be appropriate for deriving contextual definitions and requirements?
    \item[RQ 1-3:] How to express and document explicit or implicit beliefs based on the derived contextual definitions?
\end{description}
\end{tcolorbox}

\subsubsection*{Preliminary roadmap}
A first step to answering the research questions is a qualitative exploratory study. A deeper understanding of the problem in a realistic environment can be gained with data collected in interviews and focus groups. The main goal of the interviews is to get a better understanding of the challenges in defining the context for applying machine learning. The interview partners will be function developers and experts from the OEM domain, a Tier 1, or a Tier 2 automotive supplier. \par
A thematic analysis of the collected data will be performed to create a suitable model by interconnection \cite{Creswell2017}. In a focus group the findings of the study shall be validated and evaluated by the interview participants and other relevant stakeholders.

\subsection{\textbf{PA 2}: Data requirements and quality attributes of data}
Data, and especially their representation in the form of probability distributions are the core of machine learning. Different types of data (input data, training data, test data, etc.) play a role when deploying and using machine learning or deep learning. Each type can even further be categorised: For an autonomous driving system there could be driver data, vehicle data, surround data, and global data for example. The other use cases have similar data categories such as user data, sensor platform data, or cloud data.

%\begin{figure}
%    \centering
%    \includegraphics[width=0.7\linewidth]{figs/06_data_types.pdf}
%    \caption{Examples for different data used in applications with machine learning.
%    }
%    \label{fig:06_data_types}
%\end{figure}

The contexts in which a machine learning algorithms are used govern properties of the data used in design time (e.g. during the training and testing) and during runtime. Furthermore, data often originate from many different sources with varying degree of quality, which can be a challenge for the ability of ensuring the system's behaviour in a given context. Especially if system quality attributes, such as safety or robustness, must be ensured, it is crucial that the used data are trustworthy, timely, and of the required quality. An AI will be trained with data representing a context in which the system is expected to operate. However, if the context changes over time, properties of the input data to the AI change as well \cite{Willers2020}. Based on the context, there will be requirements on the input data in order to allow the AI to arrive at the right decision. An example for a data requirement could be, that the data shall represent a given probability distribution for which the AI has been trained. Only then can a machine learning model arrive at the right decision. During operation, the system might also record data that allows developers to continuously receive feedback and to implement improvements in the overall system, e.g. through retraining to correct for a (slowly) changing context.\par
Although data are probably the most important aspect of a machine learning application, there is no proper system to determine and manage the required quality and quantity of the data. Not only since the introduction of more rigid data privacy rules, such as GDPR, there is a growing pushback to the idea to "collect as much data as possible" for a machine learning application in the hope that the right data might be among them.\par

\begin{tcolorbox}[title=Research Questions: Data attributes and requirements]
\begin{description}
    \item[RQ 2-1:] What are relevant challenges of managing data quality requirements when developing large distributed systems based on deep learning?
    \item[RQ 2-2:] What constitutes a data quality framework for developing large systems based on distributed deep learning?
    \item[RQ 2-3:] To what extend can relevant challenges of managing data quality requirements be mitigated by a data quality framework for developing systems based on deep learning?
\end{description}
\end{tcolorbox}

\subsubsection*{Preliminary roadmap}
To find answers on the proposed research questions, we intend to combine literature review, data analysis, and a design science research methodology. From the automotive use case, typical data parameters such as precision, timeliness, dynamic range, and noise are collected from sensor specifications and data examples. Combined with interviews and workshops with ADAS and deep learning experts, and following the methodology for design science studies \cite{Pfeffers2006, Knauss2020}, the principal goal of the research is to devise a solution for understanding data quality requirements and dependencies between data types in distributed systems using machine learning models.

\begin{tcolorbox}[title=Literature references]
Previous research on data quality in software engineering and data quality frameworks serves as a starting point:
\begin{itemize}
    \item The significance of data quality in design, validation and implementation of software \cite{Bobrowski1998}.
    \item A proposed data quality framework for distributed computing environment \cite{Fletcher1998}.
    \item The effects of data quality on machine learning algorithms \cite{Sessions2006}.
    \item A data quality assessment and monitoring framework \cite{Batini2007}.
    \item Characteristics and challenges of big data environments \cite{Cai2015}.
    \item Reporting mechanisms of data quality in distributed networks \cite{Kahn2015}.
    \item Requirements for data quality metrics \cite{Heinricht2018}.
\end{itemize}
\end{tcolorbox}

\subsection{\textbf{PA 3}: Performance definition and monitoring}
The next step after having established a framework for defining the context of the machine learning application, and required data attributes and requirements is the definition of performance metrics, or key performance indicators (KPI), and subsequent the setup of monitoring regimes. \par
Performance definitions and monitoring of machine learning enabled systems allows to check that the system stays within its guaranteed system behaviour. In addition, {it supports development processes by providing} developers {with} feedback on how the deployed machine learning model performs "in the field". Only a continuous monitoring of the system allows for continuous integration and control of the machine learning model. \par
While it is meanwhile common practise to use machine learning models for fault detection, there are only very few efforts to fault detection in machine learning models. How can we ensure that a machine learning model is fault-free during its operation? The questions on how faults e.g. in a deep neural network can be classified and detected are not answered yet. Beyond fault detection, fault isolation and fault handling can help ensuring safety goals for systems with AI.\par
A concrete research roadmap will be derived based on the finding of the research on contextual definition and data attributes, but some general research questions about performance monitoring of machine learning models are:

\begin{tcolorbox}[title=Research Questions: Performance definition and monitoring]
\begin{description}
    \item[RQ 3-1:] What are suitable performance metrics / KPIs for trained machine learning models in a given context?
    \item[RQ 3-2:] What approaches are possible to compare / compete different trained machine learning models for a given context?
    \item[RQ 3-3:] What faults can occur in a machine learning model? Can faults in machine learning models be classified in different categories? 
\end{description}
\end{tcolorbox}

\subsection{{\textbf{PA 4}: Human Factors}}

\begin{tcolorbox}[title=Literature references]
Previous research on human factors related to automatic vehicles and AI-intense systems:
\begin{itemize}
    \item Lee et al. highlight the danger if human factors are not sufficiently considered during AV design, related to achieving sufficient safety, trust and acceptance as well to avoid the miss-use and  disuse  of the automated technology \cite{Lee2008}.
    \item Hancock's warning that human factors must be integrated in  automation design \cite{Hancock2017}.
    \item A list of socio-technical challenges \cite{Hancocka}. 
    \item A methodology based on multidisciplinary cognitive engineering (CE+) \cite{Peter}.
    \item The User Centered Ecological Interface Design (UCEID) that enables including HF considerations in the early stages of the overall system design processes \cite{Kirsten}.
\end{itemize}
\end{tcolorbox}

When building complex, AI-intense systems and products, it is important to complement a focus on internal, technical aspects of the system (e.g. the conditions and capabilities of the system in a given context) with a focus on how the intended users will interact with it in a realistic context.
For this focus, ergonomics and human factors must be taken into account. Understanding human factors is particularly important for building a system that users accept and trust in. To achieve the desired results, it is critical to consider human factors right when the concepts are developed, i.e. as part of requirement engineering. \par
However, it is challenging to integrate human factors into the development process of complex, AI-intense systems, such as automated vehicles. One reason for this is the need to shorten the time-to-market when developing new features. Hence development teams focus more on technical parts and may not possess enough human factors competence to design them according to the users' needs. \par
Since many developing organizations are transitioning to agile or continuous system development and reject the idea of comprehensive upfront requirements, development teams cannot fall back to requirements specifications for the purpose of including human factor constraints in their design decisions. \par
New advanced and AI-enabled safety features such as for example the AEB have changed the  interaction of human drivers with the their vehicle significantly. This frees up mental resources and improves the quality of driving but also affects other traffic participants and their behavior. While AI-enabled support for driving tasks have dramatically changed in just a few years, humans have not changed in the past millennium. So, while, designing such functionalities, we need to keep in mind some key elements (limitations and capabilities) from the  perspective of humans and specifically of users. 
For example, the fact that humans override or deactivate AEB functionality has become a major limitation in its ability to make traffic safer. \cite{Manchon2020}. Such scenarios must be analysed from a human factor perspective. Thus, we suggest to investigate the extent to which human factors must be considered when analysing required functionality and quality of the system and its components, in particular in relation to modern system development approaches.

\begin{tcolorbox}[title=Research Questions: Human Factors]
\begin{description}
    \item[RQ 4-1:] In what way must human factors be considered for understanding and ensuring system behaviour of AI-intense systems?
    \item[RQ 4-2:] In what way must human factors be considered for understanding and ensuring system quality attributes of AI-intense systems?
    \item[RQ 4-3:] How can HF knowledge be effectively used in modern system development approaches?
\end{description}
\end{tcolorbox}

\subsection{\textbf{X}: Cross-Cutting research problem: Integration in modern system development}

In systems development, there is a general trend towards agile, DevOps, and continuous deployment, since such approaches promise shorter time-to-market and increased responsiveness to change \cite{Gren2020}.
To achieve these goals, organisations rely on empowered, self-organised teams that take responsibility for features from inception, over design, implementation, and test, to deployment \cite{Knauss2019}. Ideally, such empowered teams allow for fast responses to change, since teams can make decisions directly. In order to facilitate such quick decisions for AI-intense systems, and to prepare for scalability, the responsibility for any activities related to our problem areas should lie with the teams to the largest possible extent.
In order to achieve this, the organisation must provide sufficient support:
\begin{itemize}
    \item Requirements information model for context, data requirements, performance metrics, and human factors.
    \item Traceability information model that allows to identify and resolve inter-team dependencies
    \item Methodological support to generate and manage knowledge about context, data requirements, performance metrics, and human factors
\end{itemize}
\begin{tcolorbox}[title=Literature references]
Previous research on integrating requirements management into modern system development approaches as a starting point:

\begin{itemize}
    \item 
    The state of the art in these areas with respect to machine learning has recently found unsatisfactory, especially in automotive use cases where ISO26262 does not well match the nature of deep neural networks \cite{Borg2019}.
    \item An overview of RE-related challenges in scaled-agile system development \cite{Kasauli2020b}.
    \item A discussion of requirements information models that support collaboration across organizational boundaries \cite{Wohlrab2019c}.
    \item Considerations of challenges related to defining traceability strategies \cite{Maeder2013} and traceability information models for modern system development approaches \cite{Maro2020,Wohlrab2019,Cleland-Huang2017,Cleland-Huang2012}.
    \item Discussions of RE practices at scale \cite{Wohlrab2019c} and boundary objects for requirements-based coordination \cite{Kasauli2020a,Wohlrab2019a}.
\end{itemize}
\end{tcolorbox}

\begin{tcolorbox}[title=Research Questions: Integration in modern system development]
\begin{description}
    \item[RQ X-1:] What are characteristics of suitable requirements information models?
    \item[RQ X-2:] What are characteristics of suitable traceability strategies?
    \item[RQ X-3:] What are characteristics of suitable methodological support? 
\end{description}
\end{tcolorbox}
% !TEX root = main.tex
\section{Conclusion and Outlook}
\label{sec:conclusion}

Distributed AI-intense systems raise a number of challenges for requirement and systems engineering. To guarantee a desired system behaviour, to guarantee system attributes such as safety, robustness and quality, and to establish process support in an organisation, this paper identified four major problem areas that need to be solved. With use cases of complex AI-enabled applications taken from different domains (industrial, automotive, and home application) this paper showed that the definition of contextual descriptions and requirements, setting data attributes and requirements, establishing performance definition and monitoring, and human factors are four key problem areas for requirement and systems engineering of AI-intense systems that need to be solved. \par
The first problem area this paper presented is the question on how to properly define, and formulate requirements on, the context in which an AI-intense system is operated. If a trained machine learning model is placed in another context, desired system behaviour and especially safety attributes can no longer be guaranteed without retraining of the model on the new context. This raises a number of research questions that are presented in this paper and a preliminary roadmap to answering them is given. \par
The second problem area is the need of proper definitions of quality attributes of, and requirements on, data that are used in the AI-intense system. This paper argued that data are the most important of systems with AI and that system quality attributes, such as safety and robustness, cannot be ensured without the ability of ensuring such properties on the data used in the system. This paper listed a number of research questions regarding data attributes and requirements and presented a preliminary research roadmap. \par
The third problem area relates to performance definition, monitoring, and human factors. Establishing requirements on the context and data for an AI-intense system raised the questions on how the fulfilment of these requirements, and the performance of the AI-intense system can be monitored. However, it was argued that a research roadmap must to be based on the findings of the research on contextual definition and data attributes. It is important to understand first what needs to be monitored, before asking how it can be measured and monitored. \par
A fourth problem area are human factors, which especially play a role in the SHAPE-IT project. In SHAPE-IT, we aim to focus on how to integrate human factor requirements in modern, large-scale AV development process to increase the vehicle safety, acceptance and trust. \par
The aim of this paper was to present a preliminary roadmap for requirement and systems engineering research in the recently launched VEDLIoT project. It is obvious that the increase use of AI will require new approaches and tools in systems engineering. To facilitate scalability and short time-to-market, we encouraged in this paper research that allows developer teams to take responsibility for the identified problem areas. It may be promising to investigate how solutions can be integrated in a git-based infrastructure \cite{Knauss2018}.
\section*{Acknowledgements}
%\eric{Thanks to Jonas B\"argman and Alessia Knauss from SHAPE-IT, as well as to all partners in VEDLIoT - we are grateful that you provided feedback where needed.}
%\eric{Specifically name use case partners?}
%\eric{Specifically name Veoneer or better yet, make another attempt to add them as authors. Without discussing this topic with them for the past year or so, we would not have been able to describe PA1-3 in this detail.}

Thanks to Jonas B\"argman for inspiration  related to PA 4 and SHAPE-IT. 
We are grateful to all VEDLIoT partners, especially in Work~Package~2 for inspiring discussions. Thank you Oliver Brunnegard from Veoneer Sweden and Alessia Knauss from ZensAct, Sweden for your feedback and proof reading.
\bibliographystyle{IEEEtran}
\bibliography{main}

\begin{thebibliography}{10}
\providecommand{\url}[1]{#1}
\csname url@samestyle\endcsname
\providecommand{\newblock}{\relax}
\providecommand{\bibinfo}[2]{#2}
\providecommand{\BIBentrySTDinterwordspacing}{\spaceskip=0pt\relax}
\providecommand{\BIBentryALTinterwordstretchfactor}{4}
\providecommand{\BIBentryALTinterwordspacing}{\spaceskip=\fontdimen2\font plus
\BIBentryALTinterwordstretchfactor\fontdimen3\font minus
  \fontdimen4\font\relax}
\providecommand{\BIBforeignlanguage}[2]{{%
\expandafter\ifx\csname l@#1\endcsname\relax
\typeout{** WARNING: IEEEtran.bst: No hyphenation pattern has been}%
\typeout{** loaded for the language `#1'. Using the pattern for}%
\typeout{** the default language instead.}%
\else
\language=\csname l@#1\endcsname
\fi
#2}}
\providecommand{\BIBdecl}{\relax}
\BIBdecl

\bibitem{Bosch2020}
\BIBentryALTinterwordspacing
J.~Bosch, H.~H. Olsson, and I.~Crnkovic, ``Engineering ai systems: A research
  agenda,'' in \emph{Artificial Intelligence Paradigms for Smart Cyber-Physical
  Systems}.\hskip 1em plus 0.5em minus 0.4em\relax IGI Global, 2020, pp. 1--19.
  [Online]. Available: \url{http://arxiv.org/abs/2001.07522}
\BIBentrySTDinterwordspacing

\bibitem{Horkoff2019}
\BIBentryALTinterwordspacing
J.~Horkoff, ``Non-functional requirements for machine learning: Challenges and
  new directions,'' in \emph{Proc. of 27th IEEE Int Requirements Eng. Conf.
  (RE)}, Jeju Island, South Korea, 2019, pp. 386--391. [Online]. Available:
  \url{https://doi.org/10.1109/RE.2019.00050}
\BIBentrySTDinterwordspacing

\bibitem{SOTIF}
ISO, ``\BIBforeignlanguage{en}{{ISO}/{PAS} 21448:2019 {Road} vehicles --
  {Safety} of the intended functionality},'' {2019}.

\bibitem{Borg2019}
\BIBentryALTinterwordspacing
M.~Borg, C.~Englund, K.~Wnuk, B.~Duran, C.~Levandowski, S.~Gao, Y.~Tan,
  H.~Kaijser, H.~L{\"o}nn, and J.~T{\"o}rnqvist, ``Safely entering the deep: A
  review of verification and validation for machine learning and a challenge
  elicitation in the automotive industry,'' \emph{Automotive Software
  Engineering}, vol.~1, no.~1, pp. 1--19, 2019. [Online]. Available:
  \url{https://doi.org/10.2991/jase.d.190131.001}
\BIBentrySTDinterwordspacing

\bibitem{Martin2017}
\BIBentryALTinterwordspacing
H.~Martin, K.~Tschabuschnig, O.~Bridal, and D.~Watzenig, ``Functional safety of
  automated driving systems: Does {ISO} 26262 meet the challenges?'' in
  \emph{Automated {Driving}}, D.~Watzenig and M.~Horn, Eds.\hskip 1em plus
  0.5em minus 0.4em\relax Springer, 2017, pp. 387--416. [Online]. Available:
  \url{https://doi.org/10.1007/978-3-319-31895-0_16}
\BIBentrySTDinterwordspacing

\bibitem{Salami2020}
\BIBentryALTinterwordspacing
B.~{Salami}, K.~{Parasyris}, and et~al., ``Legato: Low-energy, secure, and
  resilient toolset for heterogeneous computing,'' in \emph{2020 Design,
  Automation Test in Europe Conference Exhibition (DATE)}, 2020, pp. 169--174.
  [Online]. Available: \url{https://doi.org/10.23919/DATE48585.2020.9116362}
\BIBentrySTDinterwordspacing

\bibitem{j3016}
SAE, ``{SAE J3016:201806 - SURFACE VEHICLE RECOMMENDED PRACTICE - Taxonomy and
  Definitions for Terms Related to Driving Automation Systems for On-Road Motor
  Vehicles},'' {2018}.

\bibitem{Koopman2019}
\BIBentryALTinterwordspacing
P.~Koopman, U.~Ferrell, F.~Fratrik, and M.~Wagner, ``A safety standard approach
  for fully autonomous vehicles,'' in \emph{Int. Conf. on Computer Safety,
  Reliability, and Security}.\hskip 1em plus 0.5em minus 0.4em\relax Springer,
  2019, pp. 326--332. [Online]. Available:
  \url{https://doi.org/10.1007/978-3-030-26250-1_26}
\BIBentrySTDinterwordspacing

\bibitem{Gyllenhammar2020}
\BIBentryALTinterwordspacing
M.~Gyllenhammar, R.~Johansson, F.~Warg, D.~Chen, H.-M. Heyn, M.~Sanfridson,
  J.~S{\"o}derberg, A.~Thors{\'e}n, and S.~Ursing, ``Towards an operational
  design domain that supports the safety argumentation of an automated driving
  system,'' in \emph{10th European Congress on Embedded Real Time Systems (ERTS
  2020)}, 2020. [Online]. Available:
  \url{https://www.diva-portal.org/smash/get/diva2:1390550/FULLTEXT01.pdf}
\BIBentrySTDinterwordspacing

\bibitem{Mitchell2019}
\BIBentryALTinterwordspacing
M.~Mitchell, S.~Wu, A.~Zaldivar, P.~Barnes, L.~Vasserman, B.~Hutchinson,
  E.~Spitzer, I.~D. Raji, and T.~Gebru, ``Model cards for model reporting,'' in
  \emph{Proc. of the Conf. on Fairness, Accountability, and
  Transparency}.\hskip 1em plus 0.5em minus 0.4em\relax New York, NY, USA:
  Association for Computing Machinery, 2019, p. 220–229. [Online]. Available:
  \url{https://doi.org/10.1145/3287560.3287596}
\BIBentrySTDinterwordspacing

\bibitem{Wolpert1996}
\BIBentryALTinterwordspacing
D.~H. Wolpert, ``The lack of a priori distinctions between learning
  algorithms,'' \emph{Neural computation}, vol.~8, no.~7, pp. 1341--1390, 1996.
  [Online]. Available: \url{https://doi.org/10.1162/neco.1996.8.7.1341}
\BIBentrySTDinterwordspacing

\bibitem{Goodfellow2016}
\BIBentryALTinterwordspacing
I.~Goodfellow, Y.~Bengio, A.~Courville, and Y.~Bengio, \emph{The No No Free
  Lunch Theorem}.\hskip 1em plus 0.5em minus 0.4em\relax MIT press Cambridge,
  2016, ch. 5.2.1, pp. 114--116. [Online]. Available:
  \url{https://www.deeplearningbook.org/contents/ml.html}
\BIBentrySTDinterwordspacing

\bibitem{Creswell2017}
J.~W. Creswell and J.~D. Creswell, \emph{Research design: Qualitative,
  quantitative, and mixed methods approaches}.\hskip 1em plus 0.5em minus
  0.4em\relax Sage publications, 2017.

\bibitem{Willers2020}
O.~Willers, S.~Sudholt, S.~Raafatnia, and S.~Abrecht, ``Safety concerns and
  mitigation approaches regarding the use of deep learning in safety-critical
  perception tasks,'' in \emph{International Conference on Computer Safety,
  Reliability, and Security}.\hskip 1em plus 0.5em minus 0.4em\relax Springer,
  2020, pp. 336--350.

\bibitem{Pfeffers2006}
K.~Pfeffers, T.~Tuunanen, C.~E. Gengler, M.~Rossi, W.~Hui, V.~Virtanen, and
  J.~Bragge, ``The design science research process: A model for producing and
  presenting information systems research,'' in \emph{Proc. of the First Int.
  Conf. on Design Science Research in Information Systems and Technology
  (DESRIST 2006), Claremont, CA, USA}, 2006, pp. 83--106.

\bibitem{Knauss2020}
\BIBentryALTinterwordspacing
E.~Knauss, ``Constructive master's thesis work in industry: Guidelines for
  applying design science research,'' \emph{arXiv preprint arXiv:2012.04966},
  2020. [Online]. Available: \url{http://arxiv.org/abs/2012.04966}
\BIBentrySTDinterwordspacing

\bibitem{Bobrowski1998}
\BIBentryALTinterwordspacing
M.~Bobrowski, M.~Marr{\'e}, and D.~Yankelevich, ``A software engineering view
  of data quality,'' \emph{Proc. of Second Int. Conf. Software Quality in
  Europe}, 1998. [Online]. 
\BIBentrySTDinterwordspacing

\bibitem{Fletcher1998}
F.~Fletcher, ``A framework for addressing data quality in distributed computing
  systems.'' in \emph{Proc. of the 1998 Int. Conf. on Information
  Quality}.\hskip 1em plus 0.5em minus 0.4em\relax MIT Cambridge, 1998.

\bibitem{Sessions2006}
V.~Sessions and M.~Valtorta, ``The effects of data quality on machine learning
  algorithms.'' \emph{ICIQ}, vol.~6, pp. 485--498, 2006.

\bibitem{Batini2007}
\BIBentryALTinterwordspacing
C.~Batini, D.~Barone, M.~Mastrella, A.~Maurino, and C.~Ruffini, ``A framework
  and a methodology for data quality assessment and monitoring.'' in
  \emph{ICIQ}.\hskip 1em plus 0.5em minus 0.4em\relax Citeseer, 2007, pp.
  333--346. [Online]. 
\BIBentrySTDinterwordspacing

\bibitem{Cai2015}
\BIBentryALTinterwordspacing
L.~Cai and Y.~Zhu, ``The challenges of data quality and data quality assessment
  in the big data era,'' \emph{Data science journal}, vol.~14, 2015. [Online].
  Available: \url{https://doi.org/10.5334/dsj-2015-002}
\BIBentrySTDinterwordspacing

\bibitem{Kahn2015}
\BIBentryALTinterwordspacing
M.~G. Kahn, J.~S. Brown, A.~T. Chun, B.~N. Davidson, D.~Meeker, P.~B. Ryan,
  L.~M. Schilling, N.~G. Weiskopf, A.~E. Williams, and M.~N. Zozus,
  ``Transparent reporting of data quality in distributed data networks,''
  \emph{Egems}, vol.~3, no.~1, 2015. [Online]. Available:
  \url{https://doi.org/10.13063/2327-9214.1052}
\BIBentrySTDinterwordspacing

\bibitem{Heinricht2018}
\BIBentryALTinterwordspacing
B.~Heinrich, D.~Hristova, M.~Klier, A.~Schiller, and M.~Szubartowicz,
  ``Requirements for data quality metrics,'' \emph{Journal of Data and
  Information Quality (JDIQ)}, vol.~9, no.~2, pp. 1--32, 2018. [Online].
  Available: \url{https://doi.org/10.1145/3148238}
\BIBentrySTDinterwordspacing

\bibitem{Lee2008}
J.~D. Lee, ``Humans and automation: Use, misuse, disuse, abuse,'' \emph{HUMAN
  FACTORS, Vol. 50, No. 3,}, p. 404–410, 2008.

\bibitem{Hancock2017}
P.~A. Hancock, ``Imposing limits on autonomous systems,'' \emph{Ergonomics 60
  (2)}, p. 284–291, 2017.

\bibitem{Hancocka}
------, ``Some pitfalls in the promises of automated and autonomous vehicles,''
  \emph{Ergonomics :1}, 2019.

\bibitem{Peter}
P.-P. van Maanen, J.~Lindenberg, and M.~A. Neerincx, ``Integrating human
  factors and artificial intelligence in the development of human-machine
  cooperation,'' \emph{IC-AI 2005}, 2005.

\bibitem{Kirsten}
K.~Revell, P.~Langdon, M.~Bradley, I.~Politis, J.~Brown, and N.~Stanton, ``User
  centered ecological interface design (uceid):a novel method applied to the
  problem of safe and user-friendly interaction between drivers and autonomous
  vehicles,'' \emph{Intelligent Human Systems Integration,Advances in
  Intelligent Systems and Computing}, 2018.

\bibitem{Manchon2020}
\BIBentryALTinterwordspacing
J.~Manchon, M.~Bueno, and J.~Navarro, ``From manual to automated driving: how
  does trust evolve?'' \emph{Theoretical Issues in Ergonomics Science}, pp.
  1--27, 2020. [Online]. Available:
  \url{https://doi.org/10.1080/1463922X.2020.1830450}
\BIBentrySTDinterwordspacing

\bibitem{Gren2020}
\BIBentryALTinterwordspacing
L.~Gren and P.~Lenberg, ``Agility is responsiveness to change: An essential
  definition,'' in \emph{Proc. of the Evaluation and Assessment in Software
  Engineering}, 2020, pp. 348--353. [Online]. Available:
  \url{https://doi.org/10.1145/3383219.3383265}
\BIBentrySTDinterwordspacing

\bibitem{Knauss2019}
\BIBentryALTinterwordspacing
E.~Knauss, ``The missing requirements perspective in large-scale agile system
  development,'' \emph{IEEE Software}, vol.~36, no.~3, pp. 9--13, 2019.
  [Online]. Available: \url{https://doi.org/10.1109/MS.2019.2896875}
\BIBentrySTDinterwordspacing

\bibitem{Kasauli2020b}
\BIBentryALTinterwordspacing
R.~Kasauli, E.~Knauss, J.~Horkoff, G.~Liebel, and F.~G. de~Oliveira~Netoa,
  ``Requirements engineering challenges and practices in large-scale agile
  system development,'' \emph{Systems and Software}, 2020. [Online]. Available:
  \url{https://doi.org/10.1016/j.jss.2020.110851}
\BIBentrySTDinterwordspacing

\bibitem{Wohlrab2019c}
\BIBentryALTinterwordspacing
R.~Wohlrab, E.~Knauss, and P.~Pelliccione, ``Why and how to balance alignment
  and diversity of requirements engineering practices in automotive,''
  \emph{Systems and Software}, vol. 162, 2019. [Online]. Available:
  \url{https://doi.org/10.1016/j.jss.2019.110516}
\BIBentrySTDinterwordspacing

\bibitem{Maeder2013}
\BIBentryALTinterwordspacing
P.~M{\"a}der, P.~L. Jones, Y.~Zhang, and J.~Cleland-Huang, ``Strategic
  traceability for safety-critical projects,'' \emph{IEEE Software}, vol.~30,
  2013. [Online]. Available: \url{https://doi.org/10.1109/MS.2013.60}
\BIBentrySTDinterwordspacing

\bibitem{Maro2020}
S.~H. Maro, J.-P. Stegh{\"o}fer, E.~Knauss, J.~Horkoff, R.~Kasauli, J.~L.
  Korsgaard, F.~Wartenberg, N.~J. Str\o{}m, and R.~Alexandersson, ``Managing
  traceability information models: Not such a simple task after all,''
  \emph{IEEE Software}, 2020.

\bibitem{Wohlrab2019}
\BIBentryALTinterwordspacing
R.~Wohlrab, E.~Knauss, J.-P. Stegh{\"o}fer, S.~Maro, A.~Anjorin, and
  P.~Pelliccione, ``Collaborative traceability management: A multiple case
  study from the perspectives of organization, process, and culture,''
  \emph{Requirements Engineering (REEN)}, vol.~25, pp. 21--45, 2020. [Online].
  Available: \url{https://doi.org/10.1007/s00766-018-0306-1}
\BIBentrySTDinterwordspacing

\bibitem{Cleland-Huang2017}
\BIBentryALTinterwordspacing
J.~Cleland-Huang, ``Safety stories in agile development,'' \emph{IEEE
  Software}, vol.~34, no.~4, 2017. [Online]. Available:
  \url{https://doi.org/10.1109/MS.2017.108}
\BIBentrySTDinterwordspacing

\bibitem{Cleland-Huang2012}
\BIBentryALTinterwordspacing
J.~Cleland-Huang, M.~Heimdahl, J.~H. Hayes, R.~Lutz, and P.~M{\"a}der, ``Trace
  queries for safety requirements in high assurance systems,'' in \emph{Proc.
  of Int. Working Conf. on Requirements Eng.: Foundation for Software Quality
  (REFSQ)}, Essen, Germany, 2012, pp. 179--193. [Online]. Available:
  \url{https://doi.org/10.1007/978-3-642-28714-5_16}
\BIBentrySTDinterwordspacing

\bibitem{Kasauli2020a}
\BIBentryALTinterwordspacing
R.~Kasauli, R.~Wohlrab, E.~Knauss, J.-P. Steghofer, J.~Horkoff, and S.~Maro,
  ``Charting coordination needs in large-scale agile organizations with
  boundary objects and methodological islands,'' in \emph{Proc. of the Int.
  Conf. on Software and System Processes (ICSSP)}, Seoul, South Korea, 2020.
  [Online]. Available: \url{https://arxiv.org/pdf/2005.05747.pdf}
\BIBentrySTDinterwordspacing

\bibitem{Wohlrab2019a}
\BIBentryALTinterwordspacing
R.~Wohlrab, P.~Pelliccione, E.~Knauss, and M.~Larsson, ``Boundary objects and
  their use in agile systems engineering organizations,'' \emph{Journal of
  Software: Evolution and Process}, vol.~31, pp. 1--24, 2019. [Online].
  Available: \url{https://doi.org/10.1002/smr.2166}
\BIBentrySTDinterwordspacing

\bibitem{Knauss2018}
\BIBentryALTinterwordspacing
E.~Knauss, G.~Liebel, J.~Horkoff, R.~Wohlrab, R.~Kasauli, F.~Lange, and
  P.~Gildert, ``T-reqs: Tool support for managing requirements in large-scale
  agile system development,'' in \emph{2018 IEEE 26th International
  Requirements Engineering Conference}.\hskip 1em plus 0.5em minus 0.4em\relax
  IEEE, 2018, pp. 502--503. [Online]. Available:
  \url{https://arxiv.org/pdf/1805.02769.pdf}
\BIBentrySTDinterwordspacing

\end{thebibliography}

\end{document}